# A LITERATURE REVIEW ON STEWART-GOUGH PLATFORM CALIBRATIONS

**Sourabh Karmakar**
PhD Student
Mechanical Engineering
Clemson University
Clemson, SC
sourabh.karmakar@gmail.com

**Cameron J. Turner**
Associate Professor
Mechanical Engineering
Clemson University
Clemson, SC
cturne9@clemson.edu

**ABSTRACT**

*Researchers have studied Stewart-Gough platforms, also known as Gough-Stewart platforms or hexapod platforms extensively for their inherent fine control characteristics. Their studies led to the potential deployment opportunities of Stewart-Gough Platforms in many critical applications such as the medical field, engineering machines, space research, electronic chip manufacturing, automobile manufacturing, etc. Some of these applications need micro and nano-level movement control in 3D space for the motions to be precise, complicated, and repeatable; a Stewart-Gough platform fulfills these challenges smartly. For this, the platform must be more accurate than the specified application accuracy level and thus proper calibration for a parallel robot is crucial. Forward kinematics-based calibration for these hexapod machines becomes unnecessarily complex and inverse kinematics complete this task with much ease. To experiment with different calibration techniques, various calibration approaches were implemented by using external instruments, constraining one or more motions of the system, and using extra sensors for auto or self-calibration. This survey paid attention to those key methodologies, their outcome, and important details related to inverse kinematic-based parallel robot calibrations. It was observed during this study that the researchers focused on improving the accuracy of the platform position and orientation considering the errors contributed by one source or multiple sources. The error sources considered are mainly kinematic and structural, in some cases, environmental factors also are reviewed, however, those calibrations are done under no-load conditions. This study aims to review the present state of the art in this field and highlight the processes and errors considered for the calibration of Stewart-Gough platforms.*

**Keywords:** Calibration, Hexapod, Inverse kinematics, Parallel Robots

## 1 INTRODUCTION

In the world of conventional robots, there are three varieties of mechanisms: (i) Serial robot, (ii) Parallel robot, and (iii) Hybrid robot [1]. Any manipulator consists of a base and an end-effector. These are connected by multiple links. In a serial manipulator, the links are connected in series. A parallel mechanism, sometimes called Parallel Kinematic Machine (PKM) or as a Stewart-Gough or Gough-Stewart Platform is made by linking a moving body or end-effector which is generally mounted on a platform or endplate, to a reference body or base through two or more links forming a closed-loop kinematic chain [2]. The base part remains fixed. A hybrid mechanism structure is formed by combination of serial links and parallel robot. Often, the parallel robot is mounted near the end effector of the serial manipulator to provide a high precision correction to the serial system. The rigidity of a parallel robot is relatively higher than a serial manipulator. The hybrid structure increases the workspace of a parallel robot at some cost of the rigidity of the structure.

Parallel robots have received significant attention for high dynamic flexibility, structural rigidity, high accuracy due to the closed kinematic loops, no error accumulating characteristics [3], higher load-to-weight ratio, and uniform load distribution capacity compared to the serial manipulators [4]. For any parallel robots, the linking element numbers between the fixed base and movable platform vary between three to six. The link numbers together with the type of connections and the twist of the platform normally decide the degrees of freedom (DOF) of the machine.

One such parallel robots controlled by 6 links connected between the fixed base and movable platform with 6-degrees of freedom (DOF) is termed Hexapod. The hexapod was first designed by an engineer Gough from the United Kingdom in 1954 for tire testing with six actuators acting as the links between the fixed base and its moving platform. The actuators are prismatic joints. This machine had the structure of an octahedral hexapod [5]. Using the Gough's platform, in 1965 another engineer from the United Kingdom, Stewart developed an articulated 6-DOF flight simulator [6] for the training of pilots. These types of platforms are generally known as a Stewart-Gough Platform or sometimes Gough-Stewart Platform, or simply as a Stewart Platform. In this document, these terms are used to mean the same machine. The combinations of motions of the 6 actuators give the platform high precision, high structural stiffness, and high dynamic performance [7]. Stewart-Gough platforms have been employed in many fields. The potential applications of parallel robots include mining machines, walking robots, both terrestrial and space applications including areas like high-speed manipulation, material handling, motion platforms, machine tools, medical fields, planetary exploration, satellite antennas, haptic devices, vehicle suspensions, variable-geometry trusses, cable-actuated cameras, and telescope positioning & pointing devices [8]. They are used in the development of high-precision machine tools by many companies like Giddings & Lewis, Ingersoll, Hexcel, Geodetic, and others [9,10]. The application options expanded from a simulator to automobile manufacturing, inspection, human-robot collaboration, space telescope, medical tool control (by adding a hexapod at the end-effector point of a serial manipulator) [11]. For the precision and accuracy needed for these machines to perform at a specific level of operational characteristics, the platform movements must be precisely controlled. To get the necessary level of accuracy for the moving platform position and orientation, called *pose*, it is essential to understand the various errors related to the machine at the time of its operation and apply suitable compensation. Calibration of the hexapod identifies these errors and adds suitable amounts of compensations to get reliable and predictable [12] output data.

This paper surveys the calibration methods used for hexapod platforms based on Stewart-Gough platforms. Efforts were made to cover most of the key articles published after the year 2000 that are based on inverse kinematics calibration techniques. The paper has been divided into six main sections. The beginning section serves as an introduction. The second section reviews the kinematics of hexapods and the primary error factors that impact the accuracy of hexapods. Section 3 reviews the calibration techniques and the strategies used for successful calibration thereof. Major calibration methodologies and their outcomes are presented in Section 4. Section 5 discusses and compares the methods previously presented. Finally, section 6 provides a conclusion and recommendations for future work.

## 2 HEXAPOD KINEMATICS & ERROR FACTORS

A general 6-6 hexapod configuration is shown in Figure 1, with the appropriate terms defined using the nomenclature from Tsai [13] and Lee [14].

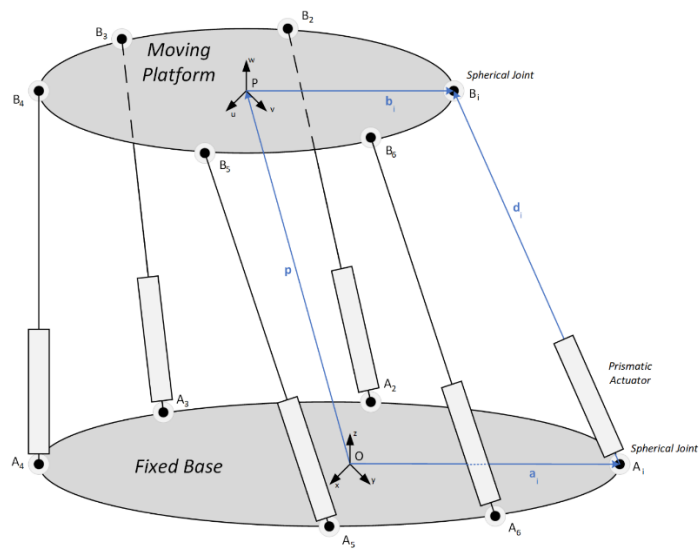

Figure 1: Hexapod Schematic with Nomenclature.

In Figure 1, the position of the Fixed Base coordinate system (System O) defined at point O compared to the Moving Platform coordinate system (System P) defined at point P is defined through the translation vector **p**, and a rotation matrix $^O_P \mathbf{R}$. The rotation matrix can be further defined as in Equation 1 as:

$$^O_P \mathbf{R} = \begin{bmatrix} u_x & v_x & w_x \\ u_y & v_y & w_y \\ u_z & v_z & w_z \end{bmatrix} \quad (1)$$

Where **u**, **v**, and **w** are unit vectors along the axes of the moving platform coordinate system. Subject to the orthogonal conditions defined in Equations 2 through 7.

$$u_x^2 + u_y^2 + u_z^2 = 1 \quad (2)$$
$$v_x^2 + v_y^2 + v_z^2 = 1 \quad (3)$$
$$w_x^2 + w_y^2 + w_z^2 = 1 \quad (4)$$
$$u_x v_x + u_y v_y + u_z v_z = 0 \quad (5)$$
$$u_x w_x + u_y w_y + u_z w_z = 0 \quad (6)$$
$$v_x w_x + v_y w_y + v_z w_z = 0 \quad (7)$$

If we define $^O\mathbf{a}_i = [a_{ix}, a_{iy}, a_{iz}]^T$ as the vector from point O to point $A_i$ (where $i = 1, 2, …, 6$) in coordinate system O, and in coordinate system P, $^P\mathbf{b}_i = [b_{ix}, b_{iy}, b_{iz}]^T$ as the vector from point P to point $B_i$ (where $i = 1, 2, …, 6$) in coordinate system P, this allows us to define a vector loop equation for the $i^{th}$ arm of the platform as in Equation 8.

$$\overline{A_i B_i} = {}^O\mathbf{p} + {}^O_P\mathbf{R}\,{}^P b_i - {}^O\mathbf{a}_i = {}^O\mathbf{d}_i \quad (8)$$

Thus, the length of the $i^{th}$ arm controlled by the prismatic actuator $i$, is defined in Equation 9 as:

$$^O d_i^2 = \left\| {}^O\mathbf{d}_i \right\| = \left[ {}^O\mathbf{p} + {}^O_P\mathbf{R}\,{}^P b_i - {}^O\mathbf{a}_i \right]^T \left[ {}^O\mathbf{p} + {}^O_P\mathbf{R}\,{}^P b_i - {}^O\mathbf{a}_i \right] = {}^O\mathbf{d}_i \bullet {}^O\mathbf{d}_i$$
$$\forall i = 1, 2, ..., 6 \quad (9)$$

Depending upon the inputs and outputs to and from the kinematic problem, the solution to the resulting system of equations is defined as either *forward kinematics* and *inverse kinematics* [15]. In forward kinematics, the position and orientation of the moving plate are calculated based on the length and orientations of the six actuators, defined by $d_i$ and expressed as $^O\mathbf{p}$ and $^O_P\mathbf{R}$.

The opposite calculation is inverse kinematics where the position and orientation of the moving plate, $^O\mathbf{p}$ and $^O_P\mathbf{R}$, are known and the required length of each of the $i$ actuators, defined by $\|\mathbf{d}_i\|$, is to be determined [16].

In both cases, the correctness of the hexapod parameters is dependent on a number of error factors which can be geometric or non-geometric [17, 18]. These error parameters affect the values of all of the variables which define the kinematics of the hexapod. Depending on the source of the different error factors, the calibration process can be classified into three levels [12,19,20]:

- *Level-1 calibration* considers only the joint errors that play a critical role in the accuracy of the robot. This is defined as "Joint Level Calibration".

- *Level-2 calibration*, also known as "Kinematic Model Calibration", takes care of the error of the kinematic parameters.

- *Level-3 calibration*, also called as "Non-kinematic Calibration" or "Dynamic Model Calibration", captures the errors of non-geometric or quasi-static parameters [21] such as stiffness, geometry of the robot structure, and errors caused by temperature variation [22].

In a hexapod platform, the components of the rigid structure like the base, frame, top platform, and other accessories are fabricated and normally made from metal stocks. So, the accuracy of these components has a direct influence on the accuracy of the whole system. The dimensions of the structure are dependent on the design tolerances or manufacturing deviations, clearances, joint errors [23], thermal deformations [24,25], and elastic deformations [26]. The actuators or struts are connected to the structure with movable joints; joints are impacted by the errors due to the assembly deviations in the form of joint run-out and ball screw deviations. Also, the mechanical joints are not free from friction, hysteresis, and backlash [3]. If the struts are operated by hydraulic fluids, there are chances of transmission errors and sensor errors [27].

Therefore, the error of a hexapod, $\mathbf{E}_{Hexapod}$, can be expressed broadly by a function of all of the terms included in Equation 10.

$$\mathbf{E}_{Hexapod} = f\left(\mathbf{E}_{manufacturing}, \mathbf{E}_{assembly}, \mathbf{E}_{transmission}, \mathbf{E}_{deformation}, \mathbf{E}_{sensor}\right) \qquad (10)$$

These error parameters are further elaborated in Table 1. Each of these errors can affect the position and orientation of the endplate. For instance, $\mathbf{E}_{manufacturing}$, is the resultant error vector due to tolerances and manufacturing accuracy along the kinematic chains that define the hexapod. Conceptually, each degree of freedom in a kinematic chain is defined by four Denavit-Hartenburg (DH) parameters, representing the information necessary to transform the coordinate system across the component providing the degree of freedom (DOF). So, on a 6-DOF kinematic chain defining one limb of a Stewert Platform, there are 4(6) or 24 DH parameters. With one prismatic actuator, there are 23 uncontrolled parameters and 1 controlled parameter in each limb. With six limbs, there are a total of 6 controlled

DH parameters and 138 uncontrolled DH parameters in a Stewert platform. However, because these parameters are coupled into closed loop kinematic chains, these parameters are coupled to the other parameters in the system, resulting in 6-independent DOF for the endplate, and the remaining 138 uncontrolled parameters are dependent upon each other and the 6-independent DOF from the prismatic actuators. Errors originating from manufacturing tolerances and accuracy can have a cumulative effect that alters all of the uncontrolled DH parameters.

Table 1: Error function illustrations

| Error source | Dependency | Remarks |
| --- | --- | --- |
| Manufacturing | Component Tolerances | These components act as the basic structure of the machine and any deviation impacts permanently. This structure bears all the loads generated in the static and dynamic condition of the machine and provides rigidity to the machine. |
| Assembly | <ul><li>Assembly Tolerances</li><li>System Age</li><li>Amount of Usage/Wear</li></ul> | Generally, assembly deviations are controllable and minimized by the replacement of old, worn-out parts with new parts. |
| Transmission | <ul><li>Actuation Response Time</li><li>Joint Clearance and Backlash</li><li>Platform Position</li><li>Operation Speed & Lag</li><li>Hysteresis Effects</li></ul> | This error depends on the robustness of the system and system configuration. Some default limitations cannot be avoided. |
| Deformation (Mechanical) | <ul><li>Material Properties</li><li>Applied Loads</li><li>Component Geometry</li><li>Platform Position</li></ul> | Dependent on the structural materials used and its response property under load. |
| Deformation (Thermal) | Working Temperature Variations | Changes in operating conditions due to temperatures may affect structural components, joint tolerances and actuator performance. |
| Sensor | <ul><li>Specification Tolerances/Accuracy</li><li>Calibration Drift</li></ul> | Modern sensors tend to be the least inaccurate. |

Similarly, assembly errors can also lead to changes in any of these parameters, although some errors may be more likely to be seen in certain parameters (i.e. wear is more likely to affect parameters specifically associated with the joint rather than the links). Similar arguments can be made for the other error components described in Table 1.

Some of the error sources, in particular the transmission and sensor errors, also can affect the controlled parameters associated with the actuators of each limb. Consequently, the task of calibrating a hexapod involves confirming that each of these 144 DH parameters is known throughout the operating range of the system.

There are more typical sources of errors [28], other than those mentioned here, contribute to the outcome from the system; however, their influences are dependent on the construction philosophy adopted for that particular system, and the type of operations and level of accuracy expected out of those systems.

## 3   ~~CALIBRATION~~ ACCURACY IMPROVEMENT STRATEGIES

The purpose of kinematic calibration of parallel robots is to improve the motion and position accuracy of the moving platform by correctly evaluating and calculating the kinematic parameters within its defined workspace. A parallel robot can be calibrated in three ways (Figure 22): External calibration, Constrained calibration, and Auto or Self-calibration [29, 63].

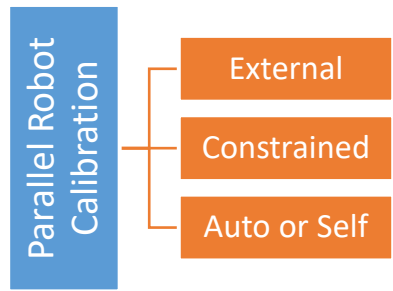

Figure 2: Strategies for parallel robot calibration.

*External calibration* is done by using one or more external instruments such as an electronic theodolite or a laser tracker [30] for measurement of multiple poses of the end-effector. In the *Constrained calibration* process, motions of the mechanical elements, usually the movement of robot actuators are constrained to gather the error data. This method is comparatively simple and the least expensive [31]. *Auto or Self-calibration* is one of the most expensive and complex calibration techniques. In this method, the robot itself automatically monitors error parameters measured by redundant sensors with the help of the built-in algorithms installed on controllers. The error correction process can take place during normal robot operation. Several extra sensors are installed in the joints and links of the robot to gather calibration data continuously. Alternatively, additional sensors can be added to directly measure the end plate motion through extensometers or feeler arms [32,33]. Generally, the number of sensors used in a parallel robot is equal to its degrees of freedom. In all these strategies, sensors play an important role and they are an essential part of the calibration process; the difference occurs on how these sensors are employed.

Conventionally, the four steps shown in Figure 33 are followed in the calibration process of a parallel robots: kinematic modeling, measurement, identification, and implementation [34–36].

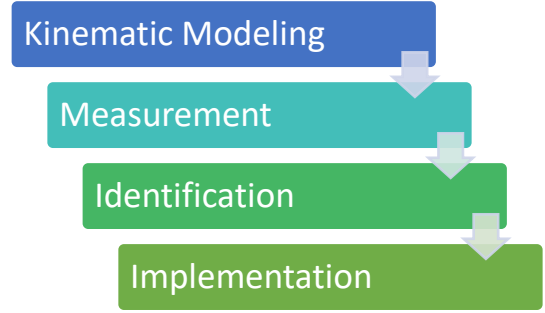

Figure 3: Steps in parallel robot calibration process.

Kinematic modeling of the platform is to build the relationship between the joint variables and the platform pose, along with the measurement device readings. The result of the modeling phase is a set of analytical equations showing relationships between these parameters as shown in Equations 1-9 [17, 18, 37]. The measurement phase gathers the data related to the actual platform position and orientation with the help of the measurement devices [21]. The identification step will identify the optimal set of unknown parameters based on the residual between the kinematic model and measurements to fit the actual behavior of the mechanism while considering the major sources of errors for the context [7, 38] and finally, through implementation, the compensation for the calculated model errors are included in the robot controller [36].

The results of a robot calibration process are expressed in terms of the pose errors for a set of positions and orientations [39]. To generate a reliable, accurate result the mechanical structure of the robot must be defined with an adequate number of parameters without repetition in the calibration model. A "good" calibration model must have three criteria: completeness, equivalence, and proportionality. Completeness refers to the fact that the model must have enough parameters to completely define the motion of the robot. Equivalence means that the derived functional model can be related to any other similar acceptable model. And the proportionality property must give the model the ability to reflect small changes in the robot geometry with small changes in the model parameters [19]. A "good" calibration model can be established by building relationships between the independent parameters that are used to define the robot system. For a multiloop parallel robot such as the 6-DOF hexapod or Stewart-Gough platform, the number of independent parameters (C) can be calculated as follows [40]:

$$C = 3R + P + S + SS + E + 6L + 6(F - 1) \tag{11}$$

where…

- L, number of independent link loops in the robot,
- R, number of unsensed (non-instrumented) revolute joints,
- P, number of prismatic joints,
- S, number of unsensed (non-instrumented) spherical joints,
- SS, number of pairs of S-joints connected by a simple link without any intermediate joint,
- E, number of measurement devices or transducers, and
- F, number of arbitrarily located frames.

For a Stewart-Gough platform with one universal (U), one prismatic (P) and one spherical (S) joints for each actuator considering [UPS ≈ 2RP3R], the minimum total number of independent parameters necessary for the complete calibration model is as follows.

$$C = 3*6*5 + 6 + 0 + 6 + 6*5 + 6(2-1) = 138$$

## 4 CALIBRATION APPROACH

To achieve a high level of kinematic accuracy, it is necessary to formulate a robust and reliable calibration method. After the introduction of hexapod by Gough, and then by Stewart, calibration became the field of interest for researchers. Calibration research is typically performed either by analytical approaches or by physical experiments. The analytical approaches are independent of the physical hexapod platform artifacts. Still, they are worth studying to get an idea of the research direction on the analytical hexapod calibration process.

### 4.1 Calibrations through Analytical Approaches

Jing et al. [41] considered that the final position and orientation of a hexapod platform is dependent on the joint radius and angles of the movable platform. In their analysis, they applied an interior-point algorithm. To validate their considerations, analytical techniques were used to reduce the errors of the 6 actuators. Through their analysis they were able to reduce the average actuator length error from 0.1 mm to 2 x 10E-7 mm.

In their research, Agheli *et al* [43] take into account the lengths of the actuators as the main sources of error for the moving platform accuracy in the hexapod and accordingly designed the calibration process to minimize the cost function though an analytical calibration process. They achieved ~50% error reduction in platform position and orientation errors at the workspace boundaries. For their analytics, Agheli *et al* used Levenberg-Marquardt algorithm to minimize the cost function obtained through inverse kinematics calculations.

In [44], the researchers Daney *et al* presented a method of calibration based on interval arithmetic and interval analysis to solve an over-constrained equation. They used Taylor expansion to obtain a linear approximation to determine the kinematic parameters. The least-square method and their developed certification algorithm provide exact kinematic parameters when no errors are considered for the measurements. With consideration of kinematic errors, the results are comparable with the classical least-square method.

For small numbers (~10) of measurements, Daney *et al* [45] used Algebraic variable elimination and monomial linearization to calibrate the platform pose. They used the actuator lengths as the error source and compared their algorithm with classical non-linear least-square methods. They obtained the same results. The advantage of this algorithm is that for small numbers of measurements, there is no need for any hypothesis for noise distribution and no initial estimate of solution is taken.

Wang *et al* [46] identified that the errors from the actuator lengths are the most dominant error factor for the overall accuracy of the 6-DOF platform. They considered the errors from the actuator lengths, ball joint location and motion errors in their analysis. Based on the sensitivity analysis of the errors, they came out with a graphical presentation of the optimal working region for their machine tool used for the experiment.

Daney *et al* [27] used the constrained optimization method and an algorithm named DETMAX for inverse kinematics applications. In this case, they studied the impact of the errors at joint positions and actuator lengths. They found that the mean error on kinematic parameters improved from 0.2705 cm to 0.0335 cm for random poses and 0.2705 cm to 0.0023 cm for selected poses.

In another study [47], Daney *et al* carried out analytical analysis using symbolic variable elimination with numerical optimization. These yielded superior results compared to the classical direct methods and were able to reduce initial pose error by up to 99% near the boundary configurations. With this study, the authors concluded that their algorithm is reliable irrespective of the robot configuration.

A summary of research on analytical-based calibration is shown in Table 2.

Table 2: Summary of analytical work on hexapod calibration

| Reference | Error/s considered | Methods | Outcome | Performance | Important points |
|---|---|---|---|---|---|
| Jing *et al* 2020 [41] | Moving platform joint radius and angles | Interior-point algorithm | Average actuator length error reduced from 0.1 mm to 2 x 10E-7 | Use of MATLAB optimization toolbox could find local optimal solution | Though other optimization methods may work better, the authors preferred this to tryout. |
| ~~Yang *et al* 2019 [42]~~ | - | ~~Dual Quaternions, Virtual power to construct the EOM (equations of motion)~~ | ~~Faster platform pose calculation~~ | ~~Computational efficiency increased by 43.45% for 6-UPS & 38.45% for 6-PUS over traditional approach.~~ | ~~• Concise parameterization of translations, and rotations~~ ~~• Avoid singularity in motion control~~ |
| Agheli *et al* 2009 [43] | Actuator length | A least-square method based on Levenberg- | Reduction in platform pose error by ~50% at the boundary | The position and orientation errors reduced 500 and | Maximum Kinematics Parameters errors were |

| Reference | Error/s considered | Methods | Outcome | Performance | Important points |
|---|---|---|---|---|---|
| | | Marquardt algorithm | | 15000 times, respectively. | observed at the boundary of the workspace. |
| Daney et al. 2004 [44] | Actuator length | Interval arithmetic and interval analysis | The new measurement method shows comparable results to the classical least-square method | The interval analysis provided numerically certified result to kinematic calibration problem | The classical least square method may not provide realistic solutions for all cases. |
| Daney et al 2004 [45] | Actuator length | Algebraic variable elimination and monomial linearization | Classical nonlinear least-squares method and this method generates exactly same results | A superior method for small numbers (=10) of measurements. | Need no hypothesis for noise distribution and no initial estimate of the solution. |
| Wang et al 2002 [46] | Actuator length, location, and motion errors of ball joints | An automated error analysis model of first and second order inverse kinematics | Graphically sensitivity analysis results to select optimal working region | 1 mm actuator length (z-axis) error causes platform deviation in x-axis from -140 to -180 µm, y-axis from 520 to 650 µm, z-axis from -150 to -350 µm | The length error (z-direction) of the actuators influences the accuracy of the machine much larger than any other errors. |
| Daney 2002 [27] | Joint position and actuator length | Constrained optimization method and DETMAX algorithm | Measurement noise in kinematic parameters reduced by 10 to 15 times | Mean error on kinematic parameters improved from 0.2705 cm to 0.0335 cm for random poses & to 0.0023 cm for selected poses. | The error value decreases steadily with an increase in the number of randomly chosen poses and remains usually constant for carefully chosen configurations. |
| Daney et al 2001 [47] | Platform pose | Symbolic variable elimination with numerical optimization | More reliable algorithm irrespective of configurations compared the standard direct methods | Initial pose error was reduced by 99%. | • An efficient technique to enhance the robustness of the measurement process. Possibility of this method for the self-calibration process. |

## 4.2 External Approach

External approaches are a widely used as a method for hexapod calibration. In this approach, the calibration is done through experiments on the hexapods using additional measuring instruments. Instruments like a double ball bar, laser interferometer, laser tracker, digital cameras, etc. are used [63]. Highlights of external approach based hexapod calibration have been summarized in Table 3 and Table 4. The system name used for the experiment, instruments used, type of error considered, and kinematics are presented in Table 3; whereas the methods, performance, and key findings of the study are presented in Table 4 for the same experiments mentioned in Table 3.

Table 3: Summary of external approaches (part 1 – left side)

| Reference | System Name | Instruments used | Error considered / measured | Kinematics |
|---|---|---|---|---|
| Song et al 2022 [48] | - | FARO measuring arm | Joint errors | Inverse |
| Mahmoodi et al 2014 [49] | - | 6 rotary sensors on 6 actuators | Actuator length | Forward/ Inverse |
| Jáuregui et al 2013 [17] | Secondary mirror of a radio-telescopes. | Laser Interferometer | Actuator length | Inverse |
| Ren et al 2009 [34] | XJ-HEXA | Biaxial Inclinometer & Laser tracker | Actuator length | Inverse |

| Reference | System Name | Instruments used | Error considered / measured | Kinematics |
|---|---|---|---|---|
| Nategh et al 2009 [50] | Hexapod table | Digital camera | Platform pose | Forward/ Inverse |
| Großmann et al 2008 [51] | FELIX | Double Ball Bar | Platform pose | Inverse |
| Liu et al 2007 [52] | - | 3D Laser Tracker | Actuator lengths | Inverse |
| Ting et al 2007 [53] | Micro-positioning platform | DMT22 Dual Sensitivity Systems with C5 probe | Hysteresis of Piezoelectric actuators | Inverse |
| Daney et al 2006 [16] | DeltaLab robot | CCD camera. | Joint imperfection and backlash of each actuator. | Inverse |
| Dallej et al 2006 [54] | - | Omni-directional camera | Position and orientation of the actuators | Inverse |
| Daney et al 2005 [55] | DeltaLab "Table of Stewart" | Sony digital video camera | Platform pose | Inverse |
| Gao et al 2003 [56] | FFCM of FAST | Laser Tracker LTD500 | Platform pose | Inverse |
| Renaud et al 2002 [57] | - | A CCD camera & 1D laser interferometry | Platform pose | Inverse |
| Week et al 2002 [58] | Ingersoll HOH600 | Double ball bar as 7th actuator & a CMM | Actuator length | Inverse |
| Ihara et al 2000 [59] | - | Telescoping magnetic ball bar (DBB) | Actuator length and joint positions | Inverse |

Table 4: Summary of external approaches (part 2 – right side)

| Reference | Methods | Performance | Important points |
|---|---|---|---|
| Song et al 2022 [48] | Artificial Neural Network (ANN) based non-linear functions | Mean pose error reduction from 0.642 mm and 0.184 deg to: Coupled network: 0.076 mm & 0.024 deg respectively; Decouple network: 0.052 mm & 0.018 deg respectively | • The coupled and decoupled networks show a similar results pattern though the optimal numbers of hidden nodes for couple network is 13 and decoupled network is ~6. |
| Mahmoodi et al 2014 [49] | A new method with 6 rotary sensors | Positional and orientation variances improved by 0.16 m$^2$ & 0.16 rad$^2$ respectively. | • The new method is less accurate in orientation measurement.<br>• This method is better than conventional method for position measurement. |
| Jáuregui et al 2013 [17] | Simplified method (same error amount considered for all actuators) & comprehensive method (each actuator error is not equal) | Majority of pose deviations fall within 10μm. | • The simplified method creates linear relationships and is easy to solve.<br>• The comprehensive method is complex, non-linear, but more accurate. |
| Ren et al 2009 [34] | Keeping any two attitude angles of the end-effector constant. | • Position accuracy = 0.1 mm<br>• Orientation accuracy = 0.011° | • Exempting the need for precise pose measurement and mechanical fixtures.<br>• Independent of inclinometer range and accuracy. |
| Nategh et al 2009 [50] | A least-square approach based on Levenberg-Marquardt algorithm with Singular value decomposition. | The position & orientation errors as per analytical methods were 0.1 mm and 0.01° respectively and were 1.45 mm and 0.27° as per experiment. | Employed Observability index to find the most visible & optimum number of measurement configurations. |
| Großmann et al 2008 [51] | Genetic Algorithm based Trajectory optimization. | The deviation of platform pose reduced from 0.7 mm to 0.17 mm. | Genetic algorithms are slow to get the most accurate solution, also rarely improve the solution. |
| Liu et al 2007 [52] | Genetic Algorithm | After 5000 generations the platform position & orientation errors improved 1.4 & 2.4 | The genetic algorithm showed good calculation stability, though it is not sensitive to measurement noises. |

| Reference | Methods | Performance | Important points |
|---|---|---|---|
| | | times respectively without measurement noise filter. | |
| Ting et al 2007 [53] | Preisach model | Platform accuracy level achieved 1 μm in position and 10 μ deg in orientation. | The convergence of errors for fixed points can happen after several iterations. |
| Daney et al 2006 [16] | Interval arithmetic and analysis methods | Yielded intervals for the position and orientation, which includes noise and robot repeatability error. | Finds ranges of parameters that satisfy the calibration model. |
| Dallej et al 2006 [54] | Linear regression | Experimental validation of the method yielded 0.8 cm median error with respect to the CAD geometry. | • An omnidirectional camera overcomes the self-occlusion problem arising in the single perspective camera.<br>• No mechanical modification of the robot is necessary. |
| Daney et al 2005 [55] | Constrained optimization method with Tabu search | Improvements in accuracy were not as per expectation due to the biasness error of 1.29 mm on the z-axis. | • The workspace boundary has a concentration of optimal poses.<br>• By maximizing the observability index, the robustness of calibration increases with respect to measurement noise. |
| Gao et al 2003 [56] | Least Square method | Accuracy improved to 0.2 mm | • Method is effective even in lack of sufficient measurements.<br>• Some false parameters may occur for fewer measurement configurations. |
| Renaud et al 2002 [57] | Error function minimization. | Precision obtained in camera measurement is in the order of 1 μm in translation and $1 \times 10^{-3}$ deg in rotations for an axial displacement of 400 mm | • Low cost and easy to use compared to the measurements by interferometer.<br>• Precision level depends on the camera resolution. |
| Week et al 2002 [58] | Gravity compensation | • Roundness accuracy improved by 3.7x<br>• Squareness accuracy improved by 7x. | The redundant actuator can be used to measure and compensate for the deflections due to gravity and thermal error. |
| Ihara et al 2000 [59] | Fourier transformation | Machine's motion error decreased to ¼. | The measurement is easy and can take care of circularity, absolute radial error, and circle center position error. |

The artificial neural network (ANN) was used by Song et al [48] in the calibration process for their Stewart-Gough platform. They corrected the joint variables by embedding the compensations in their numerical control system for online real-time error compensation. They did the experiment with their hexapod and demonstrated that the proposed ANN based robust compensator can substantially enhance static pose accuracy for both coupled and decoupled networks. The ANN approach was implemented with inverse kinematics. The results obtained show that mean pose error reduced from 0.642 mm and 0.184 deg to 0.076 mm & 0.024 deg respectively for coupled network and to 0.052 mm & 0.018 deg respectively for decouple network though the optimal numbers of hidden nodes for couple network is 13 and decoupled network is ~6.

Mahmoodi et al [49] proposed a new method of calibration for Stewart-Gough platform-based parallel robot. They used rotary sensors in place of the linear sensors in actuators. The method is not too sensitive to the orientation measurement but showed better results in position measurement for the platform. In this study, 6 rotary sensors were used on 6 actuators to correct the pose of the platform. They used a mix of forward and inverse kinematics for their platform and observed the positional and orientation variances improved over the conventional methods to 0.16 m$^2$ and 0.16 rad$^2$ respectively for both small and moderate movements.

The studies by Jáuregui *et al* [17] used a laser interferometer as the measuring instrument to calibrate the hexapod. They used inverse kinematics and considered the error related to the actuator length. Their experiment consisted of two methods. In the first method, they considered the error from all actuators to be the same, applying linear relationships. This was simple and easy to solve. In the second method, labeled it as the comprehensive method, each actuator error was measured separately and modeled in a non-linear relationship. As expected, the second method resulted in complex calculations but yielded greater accuracy.

Ren *et al* [34] did their experiment with their hexapod named XJ-HEXA using a biaxial inclinometer with the length precision of 0.002 mm and repeatability for angles of 0.001°. They reached a position and orientation accuracy up to 0.1 mm and 0.01° respectively after calibration of 80 configurations. They kept two of the three orientations defining angles of the moving platform constant during the measurements.

Nategh *et al* [50] studied their "Hexapod Table" with the use of an image capturing system. In this research, the results obtained by the analytical approaches and experiments matched very closely. The platform position & orientation errors as per the analytical methods were 0.1 mm and 0.01° respectively, whereas those from the experimental methods were 1.45 mm and 0.27°, respectively. A least-squares approach based on the Levenberg-Marquardt algorithm was employed in this calibration process with Singular value decomposition.

In another study by Großmann *et al* [51] used a Double Ball Bar (DBB) to identify and collect the kinematic parameters by moving the platform on a specific trajectory in the 3D workspace. They used a genetic algorithm and simulated measurements to finalize the parameters. Their hexapod named FELIX was designed and manufactured with a focus on simplicity and capacity for compensating the motion errors generated due to the thermal and elastic deformations. Thermal and elastic deformations were considered in the algorithm by incorporating fixed factors. By this method, they measured the error along a trajectory and were able to reduce the initial deviation from 0.7 mm to 0.17 mm after optimizing the trajectory orientation through kinematic calibration.

Liu *et al* [52] also used a genetic algorithm for calibrating their hexapod using inverse kinematics. They measured the errors coming from the actuator lengths and used a 3D laser tracker for the measurements. The genetic algorithm converged initially very fast but gradually became slow little by little. For their experiments, though they obtained an improvement in the error of the platform for the position by 1.4 times and for the orientation by 2.4 times, they found that the genetic algorithm is not sensitive to the measurement noises.

The applications of hexapod did not remain restricted to the dimension level of mm or inch, it has attracted attention for micro-level applications too. Ting *et al* [53] did their experiment with a 6-dof micro-positioning platform to evaluate the platform error due to the hysteresis of the piezoelectric actuators. By using inverse kinematics with the Preisach model, they achieved a platform accuracy at the level of 1 μm in position and 10 μ deg in orientation after several iterations. For their experiment, they used Lion Precision DMT22 Dual Sensitivity Systems with C5 Probe for measurement.

A popular method of hexapod calibration includes vision-based data collection. Daney *et al* [16] employed calibration processes using a 1024x768 CCD camera. In their experiments, they used plates with dot marks as visual targets and obtained their images for measurement analysis. The final errors measured by them were relatively large with respect to the length of the actuators and with that they concluded that the kinematic model used was not robust.

Dallej *et al* [54] used an omnidirectional camera. The measurement of the actuators had been investigated by using images from external cameras. They used an omnidirectional camera which overcomes the self-occlusion problem arising in the single perspective camera. By using the omnidirectional camera, Dallej et al. got a median error of approximately 1 cm when compared to the CAD geometry and data obtained from the camera.

Researchers Daney *et al* did several experiments with hexapod and other parallel robots. They used this work [55] on the machine DeltaLab's "Table of Stewart" involved Sony digital video camera (1024 × 768) with a 4.2 mm focal length for measuring the joint positions and actuator lengths. In this case, they used the Constrained optimization method by combining it with the Tabu search, but the results obtained were not satisfying due to the error resulting from the bias of the system along the z-axis in the range of 1.29 mm. They selected 18 and 64 random poses for analyzing the pre- and post-calibration error values.

Gao *et al* [56] carried out their study and calibration of a Five-hundred-meter Aperture Spherical Telescope (FAST) using inverse kinematics. They used a laser tracker for measurements and controlled the position and orientation of the platform with a Stewart-Gough platform-based Fine Feed Cabin Model (FFCM). In their study, they were able to achieve the desired accuracy level of 0.2 mm for the FAST. Here they had not considered any specific error factor except the final pose of the telescope.

In their research, Renaud *et al* [57] used a CCD camera and an LCD monitor to calibrate a 6-DOF parallel machine tool by using inverse kinematics. They compared the results by comparing the measurement data obtained for the same poses by a laser interferometer. The platform pose of the machine had been measured for the comparison. The final comparison showed that the precision level obtained from the optical measurement is in the order of 1 μm in translation and $1 \times 10^{-3}$ deg in rotations for an axial displacement of 400 mm.

The findings by Week *et al* [58] considered the errors on the actuator length. They used a Double Ball Bar (DBB) as a redundant actuator in their Ingersoll HOH600 robot. The accuracy of roundness and squareness in their machining tests was improved by factors of 3.7 and 7.0, respectively. The system of setup they developed could be used to measure the errors due to the thermal and gravity load deflections at the end-effector pose.

An investigation by Ihara *et al* [59] using a Telescoping magnetic ball bar (DBB) resulted in a reduction in motion error of the platform by 25% after calibration. They used Fourier transformation and included the length error of struts, and position errors of base & platform joints of the platform for optimizing the overall error of the system.

## 4.3 Constraint Approach

The constraint calibration approach has the limitation of practical applications, for this reason this calibration approach is less popular among the researchers. Constraint approach for calibration is implemented by using some mechanical constraints to restrict the motion of one or more joints in the parallel robots [30, 63]. Normally no external measuring instrument is used. The already existing sensors in the system act as the measuring sensor. The applied motion constraint causes a reduction in the degrees of freedom for the end-effector and reduction of workspace of the platform. The kinematics parameters also get reduced. The force generated due to constraining the movement may distort the structure and impact the accuracy of the calibration. As the system loses one or more degrees of freedom and the number of sensors become more than the active degrees of freedom, the calibration process may be considered as self or auto calibration process of a system with the reduced degrees of freedom.

Ryu *et al* 2001 [30] used this approach to calibrate a hexapod "Hexa Slide Mechanism (HSM)" by constraining one actuator at a time and repeating the process for all six actuators. Ryu constrained the motion of the platform by restricting the motion of one actuator of the system and the system worked as a 5-DOF system instead of the regular 6-DOF system. There was no extra sensor used, the in-built existing actuator sensors were used for taking the needful measurements. By constraining one actuator, the initial error values of the platform position 8.0E-3 m and orientation 1.4E-2 rad converge to 3.8E-16 m & 1.7E-15 radians, respectively.

Rauf *et al* 2001 [60] constrained 3 actuators and experimented with 3-DOF in the same hexapod "Hexa Slide Mechanism (HSM)" mentioned in previous paragraph. Here also the final correction values are near zero. For 3-DOF measurements, initial error values of position 1.0E-2 m and orientation 1.7E-2 radians changed to 2.4E-10 m & 1.8E-9 radians, respectively. The final correction values change by very small amounts depending on the value of the measurement noises. The advantage of these methods is that the locking device can be universal and need not be specific for a particular system.

### 4.4 Auto or Self Calibration Approach

Auto or Self-calibration is one of the ways to calibrate 6-DOF hexapod platforms. This method requires adding one or more redundant sensors to the passive joints or an additional redundant passive limb [63]. The addition of extra sensors increases the complexity of the design and manufacturing processes of platforms. Moreover, the addition of redundant sensors sometimes makes system development more expensive. The auto or self-calibration method also limits the workspace for calibration. Some research studies have been done with this method.

Mura [32] used a set of wire extensometers to directly measure the position of the moving end plate relative to the fixed plate during testing of flexible automotive components. The testing motions used ensured that the extensometers remained in tension throughout the test and provided accurate positional data while not imparting a significant amount of additional stiffness into the system.

Similarly, Guo *et al* [33] used a smaller set of four (4) extensometers to directly measure the position of the end plate relative the fixed plate during multimodal loading tests. The data was used in real-time to connect to an ADAMS model of the system, and to generate force feedback information as part of the system control loop. This use of continuous calibration demonstrated that forces could be very precisely controlled during testing.

In their study, Chiu *et al* [61] used a cylindrical gauge block and a commercial trigger probe to do the auto-calibration of the hexapod platform. They made use of the non-linear least square method in their algorithm. The advantage of their method is that the instruments are standard and commercially developed, which makes them easily available and less expensive. Multiple robot configurations were used to validate their method and the results showed that different levels of accuracy were achieved.

Similarly, in another auto-calibration process, Zhuang *et al* [62] used a Coordinate Measuring Machine (CMM) with their robot FAU Stewart-Gough Platform. Here also the position and orientation of the platform were used to calibrate system error. They used Levenberg-Marquart algorithm for optimization and got error reduction by 50%.

The highlights from their research are that some extra sensors needed to be installed in some of the joints to gather calibration data. Another conclusion they drew is that if the end-effector is a separate attachment on the hexapod platform, then the end effector should be calibrated separately.

The research done by Patel *et al* [9], observed that for their calibration algorithm increased error of the platform poses in some cases and for around 90% of observations, they got accuracy improvement from 50% to 100%. They used the least square method for their calibration algorithm. The advantage of their method is that the extra sensor is easily mountable and un-mountable; also, the extra sensor consisting of Ball-Bar can remain with the system during the actual machine operation to allow online calibration.

The details are presented in Table 5. It should be noted that for each case, platform pose error was considered.

Table 5: Assessment of auto or self-calibration approaches

| Reference | Robot Name | Instruments used | Methods | Performance | Important points |
|---|---|---|---|---|---|
| Mura 2011 [32] | - | Wire extensometer (6) | Direct Kinematics | Used to measure flexible automobile components in dynamic loading conditions such as fatigue situations. | Extensometers continuously measure the platform position and represent a level of stiffness that is negligible compared to the measurement item. |
| Guo *et al* 2016 [33] | MLMTM | Wire extensometer (4) | Direct Kinematics & ADAMS Simulations | Used to measure flexible specimens in multiaxial loading situations. Coupled with an ADAMS simulations to provide PID Control information for Force Control. | Integrated into the control loop to provide deflection information to an ADAMS model to provide data to a PID force control loop. The extensometers represent a negligible stiffness in comparison to the test specimens. |
| Chiu *et al* 2003 [61] | - | A cylindrical gauge block and a commercial trigger probe | Nonlinear least squares | Multiple platform configurations were used to validate the calibration process and different accuracy levels have been obtained. | The instruments used here are standardized and commercialized. The method is comparatively compact and economical. |
| Zhuang *et al* 2000 [62] | FAU Stewart-Gough Platform | CMM | Levenberg-Marquart algorithm | The average error was reduced by more than 50 percent. | The end-effector required separate calibration since it was not part of the closed-loop kinematic chains. It requires redundant sensors that need to be installed at some of the joints of the machine tool. |
| Patel *et al* 2000 [9] | - | Ball-bar | Least square minimization | Analysis suggests that the measuring device accuracy needs to be five to ten times more than the desired calibration accuracy. | The extra actuators can be mounted or unmounted easily. When left with the machine in certain situations, it enables online calibration. |

## 5   DISCUSSION AND COMPARISON

The goal of each calibration method is to make the hexapod machines more accurate, improve precision and obtain correct results at the platform pose during their operations. Based on the literature survey done, it can be inferred that researchers predominantly opt for conducting experiments on their hexapod platforms to validate their calibration procedures. Among these experiments, the majority employed external measurement equipment. The key focus in these studies is the improvement of pose accuracy, primarily targeting platform pose errors, joint errors, and actuator length errors. The extent of error being addressed varies among these studies. In Table 6, the crucial details from each research endeavor were summarized, aiding readers in identifying literature of interest based on the chosen approach, error considerations, methods employed, and the level of error addressed.

It's worth noting that the units used to measure errors differ across various studies, and some literature sources do not specify pose errors before the calibration process; they only report final values after calibration. Additionally, in certain instances, pose error is defined solely by position errors, with no consideration given to orientation errors. Remarkably, none of the experiments focus exclusively on improving accuracy through orientation-related factors.

Table 6: Comparison of experimental calibrations mentioned in this review.

| Reference | Calibration approach | Instruments used | Error considered | Methods | Pose Error before calibration | Pose error after calibration |
|---|---|---|---|---|---|---|
| Song et al 2022 [48] | External | FARO measuring arm | Joint errors | Artificial Neural Network (ANN) based non-linear functions | 0.1 mm & 0.1 deg | 0.1 mm & 0.1 deg |
| Mahmoodi et al 2014 [49] | External | Rotary sensors | Actuator length | A new method | 0.1 m & 0.1 rad | Reduced by 8-16 times |
| Jáuregui et al 2013 [17] | External | Laser Interferometer | Actuator length | Two different methods based on error propagation calculation | 0 – 60 μm | 0 – 10 μm |
| Ren et al 2009 [34] | External | Biaxial Inclinometer & Laser tracker | Actuator length | A new orientation constraint method | 9 mm & 1 deg | 0.1 mm & 0.01 deg |
| Nategh et al 2009 [50] | External | Digital camera | Platform pose | Least-square approach based on Levenberg-Marquardt algorithm | 2.67 mm & 4.6 deg | 1.45 mm & 0.27 deg |
| Großmann et al 2008 [51] | External | Double Ball Bar | Platform pose | Genetic Algorithm based Trajectory optimization. | 0.7 mm | 0.17 mm |
| Liu et al 2007 [52] | External | 3D Laser Tracker | Actuator lengths | Genetic Algorithm | 0.7237 mm & 0.1346 deg | 0.1555 mm & 0.0172 deg |
| Ting et al 2007 [53] | External | DMT22 Dual Sensitivity Systems with C5 probe | Actuator lengths | Preisach model | - | 1 μm & 10 μdeg |
| Daney et al 2006 [16] | External | CCD camera. | Joint errors | Interval arithmetic and analysis methods | 1.32 mm & 0.34 deg | 1.10 mm & 0.26 deg |
| Dallej et al 2006 [54] | External | Omni-directional camera | Joint errors | Linear regression | 1.2 cm | ~1 cm |
| Daney et al 2005 [55] | External | Sony digital video camera | Platform pose | Constrained optimization method with Tabu search | - | 1.45 mm & 0.27 deg |
| Gao et al 2003 [56] | External | Laser Tracker LTD500 | Platform pose | Least Square method | ±2 mm | ±1 mm |

| Reference | Calibration approach | Instruments used | Error considered | Methods | Pose Error before calibration | Pose error after calibration |
|---|---|---|---|---|---|---|
| Renaud *et al* 2002 [57] | External | A CCD camera & 1D laser interferometry | Platform pose | Error function minimization. | - | 1 μm & 0.001 deg |
| Week *et al* 2002 [58] | External | Double ball bar & CMM | Actuator length | Gravity compensation | - | Improvement (μm): Roundness 3.7x Squareness 7x |
| Ihara *et al* 2000 [59] | External | Telescoping magnetic ball bar (DBB) | Actuator length and joint positions | Fourier transformation | Circle center error (-68, 164) μm | Circle center error (-7, -9) μm |
| Chiu *et al* 2003 [61] | Self-calibration | Cylindrical gauge block & trigger probe | Platform joints | Nonlinear least squares | ±1 μm | ±0.001 μm |
| Zhuang *et al* 2000 [62] | Self-calibration | CMM | Platform Joints | Levenberg-Marquart algorithm | 1 μinch | 0.1 μinch |
| Patel *et al* 2000 [9] | Self-calibration | Ball Bar | Platform Pose | Least square minimization | ±0.1 mm | 3% to 100% |
| Mura 2011 [32] | Self-calibration | Wire Extensometers | Platform Pose | Direct Kinematics | ±0.1 mm | ±0.005 mm |
| Guo *et al* 2016 [33] | Self-calibration | Wire Extensometers | Platform Pose | Direct Kinematics | ±10 N, ±0.1 Nm | ±0.1 N, ±0.005 Nm |

As mentioned earlier, the pose accuracy is dependent on several mechanical and surrounding factors like temperature and load being experienced by the system. The ideal calibration would consider all these factors for any working condition of the machine but achieving that is not only expensive and time-consuming but also potentially unnecessary depending on the application conditions. In experiments where only the actuator lengths have been considered as the sources of error for the accuracy of the platform pose, it should be noted that the motion of the platform is dependent on all the joints which are moving to generate the motion. So, while calibration may include only the actuator length error, it also indirectly includes the error contributed by the joints. Even if all joints are not equipped with individual sensors, their error factors are indirectly accounted for in the calibration process. In general, when the platform pose accuracy is of primary interest, the calibration of it indirectly considers all the error factors existing in the hexapod system but adding appropriate compensation for each error factor becomes difficult unless they are correctly identified and accounted for in the calculations.

### 5.1 Error Identification

From the tables above, it is seen that the level of improvements obtained are varied, with cases showing an increase in the error values. The error factors considered are also different in each study. In most cases, the platform pose error due to the actuator length errors remained common and was considered as the primary error contributor in the whole system. Also, considering the error of each actuator separately and equally impacts the overall accuracy of the platform. In all these calibration processes, use of external instruments is the most common practice. External instruments such as Double Ball Bar (DBB), laser interferometer, biaxial inclinometer, laser tracker, telescopic magnetic ball bar, and wire extensometers were used. A separate study may be useful to ascertain the effectiveness of each of these instruments for the calibration of a hexapod by inverse kinematics.

Optical calibration methods have also gained importance in recent times. CCD camera, omnidirectional camera, digital video camera, and other image capturing devices were used to achieve an accuracy level in the range of 1.0 cm to 0.1 mm. The advantage of using optical methods is that the parallel machine does not need modifications to accommodate the measurement equipment for calibration, also it is non-invasive and automatically records the event for future reference. In these cases, the quality of the optical systems and the associated analysis system configuration have a major contribution to the final accuracy and precision attained.

## 5.2  Algorithms

Among the several different algorithms used for the calibration purpose, the least-square method based on the Levenberg-Marquardt algorithm was used most for both optical and non-optical calibration. The genetic algorithm had been used for a couple of studies, but they appeared to have high running time to reach an optimized level and were not sensitive to measuring noises. Constrained optimization method and Tabu search, Algebraic Elimination, non-linear least square methods were used and all of them resulted in various levels of calibration accuracy. The use of quaternions has improved calculation efficiency, though the practical application of hexapod accuracy level achieved by this technique has yet to be evaluated. Different algorithms also yielded different results on the same system for the calibrations done by Daney *et al* [45] Therefore, the selection of calibration methods and algorithms plays an important role in obtaining the desired accuracy level for the parallel robot system and application. In all cases, the final platform pose is the guiding parameter to evaluate calibration outcome.

## 5.3  External factors

There are some inherent dimensional errors in the hexapod structure due to the dimensional tolerances of each component used in the fabrication. The cumulative effects of all these tolerances play a significant role in platform accuracy. Apart from these errors, other factors, hysteresis for instance, vary with the change of the operational characteristics. Normally, the effect of temperature is expected to be minimal for the parallel robot system unless the system experiences large temperature variations during its operation. From a practical point of view, such robot systems operate in a controlled environment unless they are deployed in special applications like large field telescope mounting. In these applications where exposure to fluctuating weather conditions is unavoidable, proper calibration factors for thermal deviation must be included.

Likewise, the elastic deformation error factor is not dominant in all cases. If the hexapod platform is subjected to high loads relative to its structure, a factor for elastic deflection is essential. Hexapod platforms reviewed in this paper can carry loads up to 2000 kg though none of them were subjected to such load during the calibration process. As such, the structure can undergo a substantial amount of elastic stress and that may lead to a notable amount of deflection to impact the accuracy of the robot. In these types of cases, the factor of elastic load cannot be ignored. There are potential research opportunities for evaluating the impact of load conditions on the parallel robots and to add suitable compensation factors for further improving the platform pose accuracy.

# 6    CONCLUSIONS

Hexapods are used for precise, complex, and repeatable operations in a variety of applications. Small errors in the system can lead to a serious impact on pose accuracy. Therefore, calibration is a critical activity for any hexapod machine to be reliable in each application. There are several sources of errors that negatively impact the accuracy of the hexapod. Selecting an appropriate calibration approach is dependent upon the purpose and constraints of the hexapod system. In some cases, external sensor calibration may be the best solution, while in other cases, constrained or auto-calibration may be a better approach. This paper attempts to provide an overview of those methods and draw an outline of the current state of the art in this field and help other researchers to take appropriate notes for the desired methods.

Researchers have explored a variety of calibration methods, each exhibiting varying degrees of accuracy and complexity in implementation. The focus of this review is to identify the methods used to improve the positional accuracy of the hexapod platform or end-effector and accuracy improvement obtained after calibration. Instruments and techniques utilized to compensate pose errors stemming from various inaccuracies were summarized. Individual researchers searching for an appropriate calibration approach should be cognizant of the level of calibration needed for a particular application and the resulting impact upon the complexity of implementation. Three types of calibration strategies are identified for physical systems: external, constraint, and auto- or self-calibration. Double ball bars, laser trackers, laser interferometers, and optical devices are some of the most used instruments for collecting positional data. The types of errors typically considered in these studies revolve around platform pose, joint variables and actuator length. In addition to the studies on physical systems, analytical-based studies were also discussed, and their specifics were summarized. In general, the authors agree with the sentiment of Merlet [63], who suggested that the emergence of image based systems offers a great deal of potential improvement. Since this statement was published in 2006, we have only observed improvements in the hardware and software supporting image-based calibration and we also note that biological systems generally use manipulator-image-coordination (i.e. hand-eye-coordination) approaches with great success.

The literature reviewed in this paper predominantly covers research conducted since the year 2000 focusing hexapod platforms and on calibration of platform-pose under no-load condition; however, in actual applications, these hexapods may be experiencing heavy working loads. The effect of working load propagates directly to the hexapod structure. Based on structural rigidity, the load causes elastic deformation which can affect the operational accuracy. Therefore, the response behavior and platform accuracy of a hexapod under the influence of working loads remains a subject for further studies.